\documentclass{amia}
\usepackage{graphicx}
\usepackage[labelfont=bf]{caption}
\usepackage[superscript,nomove]{cite}
\usepackage{color}
\usepackage{subcaption}

\begin{document}

\title{Using Latent Class Analysis to Identify ARDS Sub-phenotypes for Enhanced Machine Learning Predictive Performance}

\author{Tony Wang PhD$^{1}$, Tim Tschampel$^{2}$, Emilia Apostolova, PhD$^{3}$, Tom Velez, PhD, JD$^{2}$}

\institutes{
    $^1$Imedacs, Ann Arbor, MI; $^2$Computer Technology Associates, Ridgecrest, CA; $^3$Language.ai, Chicago, IL \\
}

\maketitle

\noindent{\bf Abstract}

\textit{In this work, we utilize Machine Learning for early recognition of  patients at high risk of acute respiratory distress syndrome (ARDS), which is critical for successful prevention strategies for this devastating syndrome. The difficulty in early ARDS recognition stems from its complex and heterogenous nature. In this study, we integrate knowledge of the heterogeneity of ARDS patients into predictive model building. Using MIMIC-III data, we first apply latent class analysis (LCA) to identify homogeneous sub-groups in the ARDS population, and then build predictive models on the partitioned data. The results indicate that significantly improved performances of prediction can be obtained for two of the three identified sub-phenotypes of ARDS. Experiments suggests that identifying sub-phenotypes is beneficial for building predictive model for ARDS.}

\section{Introduction}

The acute respiratory distress syndrome (ARDS) is a significant cause of morbidity and mortality in the USA and worldwide \cite{bellani2016epidemiology,pham2017fifty,bice2013cost, webster1988adult}. While the mainstay of treatment is to treat the inciting cause of ARDS effectively, it is clear that early, evidence-based management of ARDS can limit the propagation of lung injury and significantly improve patient outcomes.  Early management requires early identification.  In 2012 an expert panel announced a new definition of ARDS as the acute (within 7 days of a precipitant cause such as sepsis, trauma, etc.) development of abnormal oxygenation and bilateral opacities in chest imaging consistent with pulmonary edema. However the reported ability to predict mortality using this definition was at best modest with an AUC of 0.577 \cite{force2012acute}.   

To date, there exists no reliable way to anticipate which patients are likely to develop ARDS.  Numerous prediction scores have been developed to assess ARDS prognosis and risk of death, such as the Lung Injury Score (LIS) \cite{murray1988expanded}, Lung Injury Prediction Score (LIPS) \cite{gajic2011early}, APPS (age, plateau, PaO2/FiO2) score \cite{villar2016age}, Early Acute Lung Injury (EALI) \cite{levitt2013early}, and Modified ARDS Prediction Score (MAPS) \cite{xie2018modified}. However, the predictive validities of these ARDS scoring tools have been shown to be moderate: for example LIPS discriminated patients who developed Acute Lung Injury (ALI) from those who did not with an AUC of 0.80 (95\% CI, 0.77–0.84)7.  MAPS was shown to have a similar AUC of 0.79 (95\% CI, 0.72 - 0.87) in predicting ARDS development10; the reported EALI AUC was 0.85 (95\% CI: 0.80-0.91, (on the training set) for identifying patients who progressed to acute lung injury \cite{levitt2013early}. 

In identifying mortality, the predictive validity of LIS was found to be limited, with an AUC of 0.60 (95\% CI 0.55 to 0.65), in the era of the Berlin definition \cite{kangelaris2014there}. Similarly APPS was found to have an AUC of 0.8 for predicting ARDS mortality \cite{villar2016age}. General illness severity scores, such as SAPS, SAPS II, APACHE II/III, and MPM, were also examined in their ability in helping recognize ARDS early, but only similar moderate performance has been reported \cite{villar2016age, levitt2013early,kangelaris2014there}.

Improved predictive validity is needed to enable reliable early identification and management of patients at risk for ARDS.   We hypothesize that a barrier to improved predictive performance of existing scoring tools may be the heterogeneity of the ARDS populations used to derive these models.   Studies have indicated that ARDS is a highly heterogeneous syndrome that may be composed of several distinct sub-phenotypes \cite{calfee2014subphenotypes, sinha2018latent, zhang2018identification}. Such heterogeneity in population implies heterogeneity in relationships between explanatory and response variables within partitions, posing serious challenges in predictive model building seeking to identify a common explanatory data pattern associated with an outome \cite{karpatne2014predictive}.

In this study, we integrate prior knowledge of the heterogeneity in ARDS population into predictive model building by identifying ARDS subtypes that share common underlying pathophysiology as statistically expressed in observed clinical data (e.g. labs, vitals). Specifically we utilize latent class analysis (LCA) \cite{mccutcheon1987latent} to identify homogeneous sub-groups of ARDS subjects \cite{zhang2018identification}, and then build predictive models on partitioned data \cite{karpatne2014predictive} to see whether predictive validity could be improved, comparing with those from treating all patients as a whole homogeneous group. Prior work has demonstrated the value of problem domain semantics in the enhancement of mortality risk predictive models \cite{wang2018semantically}.

\section{Material and Methods}
\subsection{Dataset}

Clinical encounter data of adult patients (age $\geq$18 years) were extracted from the MIMIC- III version 1.4 ICU database \cite{johnson2016mimic}. MIMIC-III consists of retrospective ICU encounter data of patients admitted into Beth Israel Deaconess Medical Center from 2001 to 2012. Included ICUs are medical, surgical, trauma-surgical, coronary, cardiac surgery recovery, and medical/surgical care units. Although MIMIC-III includes both time series data recorded in the EMR during encounters (e.g. vital signs/diagnostic laboratory results, free text nursing notes/radiology reports, medications, discharge summaries, treatments, etc.) as well as high- resolution physiological data (time series / waveforms) recorded by bedside monitors, only the time series data recorded in the EMR was used in this study.

\subsubsection{Data for Latent Class Analysis}

ICD-9 diagnosis codes and procedure codes identifying mechanically ventilated patients are the basis for identifying the class of ARDS patients. PaO2, FiO2, and PEEP information were extracted from charted data. Time points of ARDS onset are defined based on Berlin criteria\cite{force2012acute}, i.e. PaO2/FiO2 ratio $\leq$ 300 with PEEP at least 5 cmH2O. The observed vital and lab measurements after the identified diagnosis time are extracted, and features constructed as class-defining variables in the LCA modelling including BMI, means of bicarbonate, plateau pressure, mean airway pressure (MAP), PaCO2, tidal volume, platelet count, total bilirubin; minimum of sodium, glucose, albumin, hematocrit, systolic blood pressure (SBP); maximum of temperature, heart rate, white blood cell (WBC) count, creatinine; and first-available PaO2/FiO2 ratio and PEEP (Table \ref{table-1}). Four predisposition conditions: sepsis, shock, aspiration, and pneumonia, are also included in the LCA analysis. 

\subsubsection{Data for Predictive Modeling} Features considered in the predictive model building include: 1) vital signs: heart rate, respiratory rate, body temperature, systolic blood pressure, diastolic blood pressure, mean arterial pressure, oxygen saturation, tidal volume; 2) laboratory tests: white blood cell count, bands, hemoglobin, hematocrit, lactate, creatinine, bicarbonate, pH, PT, INR, BUN, blood gas measurements (partial pressure of arterial oxygen, fraction of inspired oxygen, and partial pressure of arterial carbon dioxide); 4) motor, verbal, and eye sub-score of Glasgow Coma Scale ; and 5) indicators of predisposition factors and potential modifier: gender, sepsis, shock, trauma, pneumonia, high risk surgery, near drowning, fracture, diabetes. See Appendix for a complete list of features with considered time ranges and properties of features. 

\subsection{Methods}

\subsubsection{Latent Class Analysis}

Latent class model estimation is based on Gaussian finite mixture modelling methods \cite{fraley2002}. It assumes that the population is composed of a finite number of components.  Mixture model parameters, i.e. components’ means, covariance structure, and mixing weights, are obtained via the expectation maximization (EM) algorithm. Before LCA modeling, Yeo-Johnson power transformation \cite{yeo2000new} is applied to ensure approximate normality of continuous variables, and mean imputation is used to replace missing values. To select a model fitting the data best, a series of latent class models with different number of components are fitted, and Bayesian Information Criteria (BIC) is used for model selection. 

\subsubsection{Predictive Modeling}

Predictive models, including gradient boosted machine (GBM)\cite{friedman2001greedy} and random forest (RF)\cite{breiman2001random}, were built for all cases and each phenotype separately, with all non ARDS subjects as the contrast group. Missing values were replaced by medians of each variables. Synthetic Minority Over-Sampling Technique (SMOTE) is applied to resample data \cite{chawla2002smote}. Data was split into a training (70\%) and test set (30\%). Cross validation was used for tuning hyperparameters: number of trees, interaction depth, learning rate, minimum number of observations in nodes for GBM model; and number of trees for RF model.  Tuned models were used to evaluate performance of predicting sepsis in the test set.  Confidence intervals of performance metrics were obtained by bootstrapping method \cite{carpenter2000bootstrap}. To compare performance (e.g. AUC) between models, the method proposed by Delong et al. \cite{delong1988comparing} was applied, with the null hypothesis that the true difference in performance metrics was equal to 0.

\section{Results}

\subsection{Latent Class Analysis to Identify ARDS Sub-phenotypes}

Data of 4181 ARDS patients (4714 ICU stays) were used in the LCA analysis classification was conducted without consideration of clinical outcomes. Details on clinical variable selection, data cleaning and a complete list of the clinical variables included in the LCA models are listed in Table \ref{table-1}. Unless specified, medians of measurements are extracted. When value is missing values, median imputation is applied to each variable.

\begin{table}[t!]

\begin{center}
\begin{tabular}{l|l|l|l|l|l|l}
\hline
 \textbf{Variable} & \textbf{N missing} & \textbf{Mean} & \textbf{SD} & \textbf{Minimum} & \textbf{Median} & \textbf{Maximum}\\
\hline 

Albumin** & 1469 & 2.48 & 0.61 & 1 & 2.5 & 4.8 \\
 Bicarbonate** & 99 & 24.86 & 5.32 & 6 & 25 & 46 \\
 Total Bilirubin** & 1266 & 2.65 & 5.74 & 0.1 & 0.7 & 55.2 \\
 BMI & 3287 & 42.48 & 261.26 & 5.51 & 27.82 & 7508.67 \\
 Highest Creatinine & 97 & 1.02 & 0.99 & 0.1 & 0.7 & 9.9 \\
 Lowest Glucose & 95 & 85.90 & 36.78 & 0.116 & 83 & 643 \\
 Lowest Hematocrit & 92 & 25.81 & 5.18 & 2.1 & 25.1 & 55.5 \\
 Maximum Heart Rate & 85 & 124.34 & 23.89 & 55 & 123 & 239 \\
 MAP** & 307 & 10.89 & 4.09 & 2 & 10 & 41 \\
 PaCO2** & 90 & 41.22 & 9.19 & 15 & 40 & 111.5 \\
 PaO2/FiO2 ratio* & 0 & 253.94 & 192.23 & 21 & 225 & 5550 \\
 PEEP* & 194 & 6.30 & 2.64 & 0.5 & 5 & 28 \\
 Plateau Pressure** & 886 & 21.10 & 5.70 & 3 & 20 & 75 \\
 Platelet count** & 113 & 236.95 & 144.21 & 9 & 218.5 & 1328 \\
 Lowest SBP & 88 & 74.55 & 21.86 & 1 & 77 & 162 \\
 Lowest Sodium & 75 & 134.39 & 5.90 & 3.5 & 135 & 165 \\
 Maximum Temperature & 136 & 38.29 & 0.97 & 32.83 & 38.28 & 42.78 \\
 Tidal Volume** & 435 & 505.28 & 93.08 & 1 & 500 & 999 \\
 Peak WBC & 116 & 8.25 & 10.10 & 0.1 & 7 & 471.7 \\
 
\hline
\end{tabular}
\end{center}
\caption{\label{table-1} Basic statistics of variables used in LCA, first available datapoints after ARDS onset. **: Mean of observed values is extracted; *: First available measurement.}
\end{table}

\begin{figure}[h!]
\centering
\includegraphics[scale=0.65]{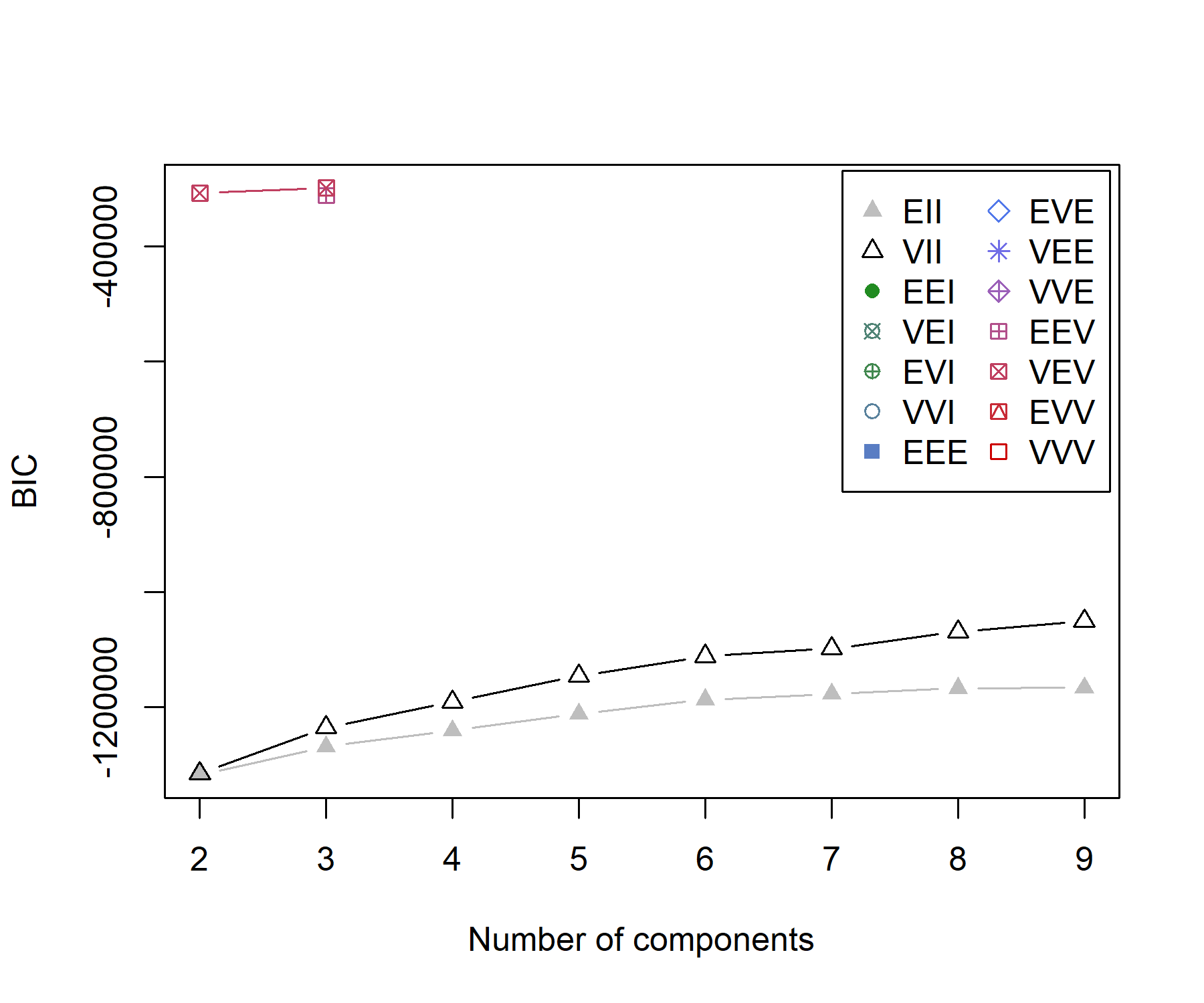}
\caption{Number of classes and corresponding BIC.}
\label{fig-1}
\end{figure}

BIC criterion suggests a 3-component VEV model (i.e. ellipsoidal distribution, variable volume, equal shape, and variable orientation) fits the data best (Figure \ref{fig-1}).  Numbers of ICU stays assigned into sub-phenotypes are 753 in phenotype 1, 1471 in phenotype 2, 2490 in phenotype 3. Figure 2 shows differences in variables by phenotype assignment. Figure \ref{fig-3} below shows mortality of each phenotype.

\begin{figure}[h!]
\centering
\includegraphics[scale=0.75]{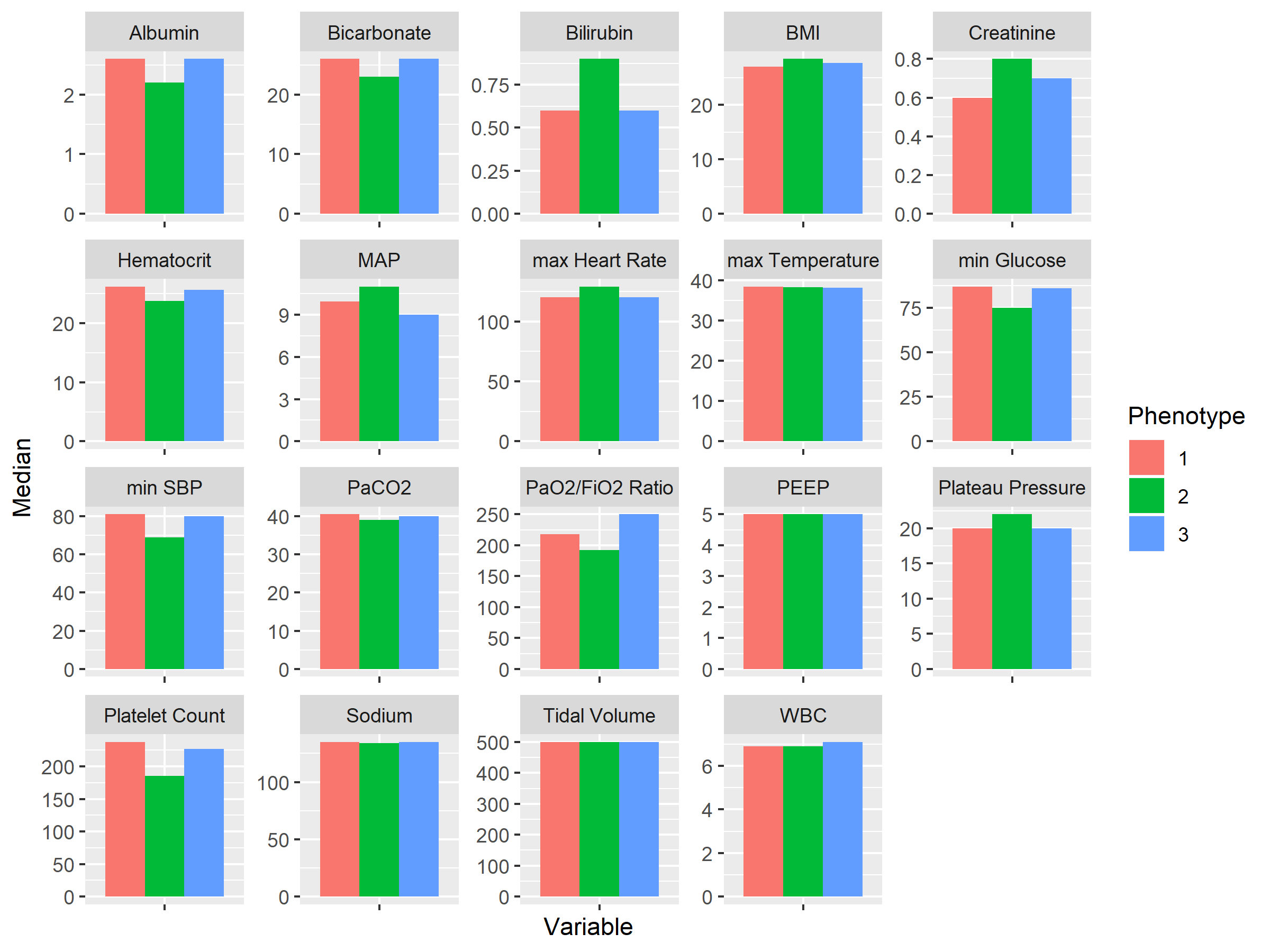}
\caption{Differences in variables by phenotype assignment.}
\label{fig-2}
\end{figure}

\begin{figure}[h!]
\centering
\includegraphics[scale=0.65]{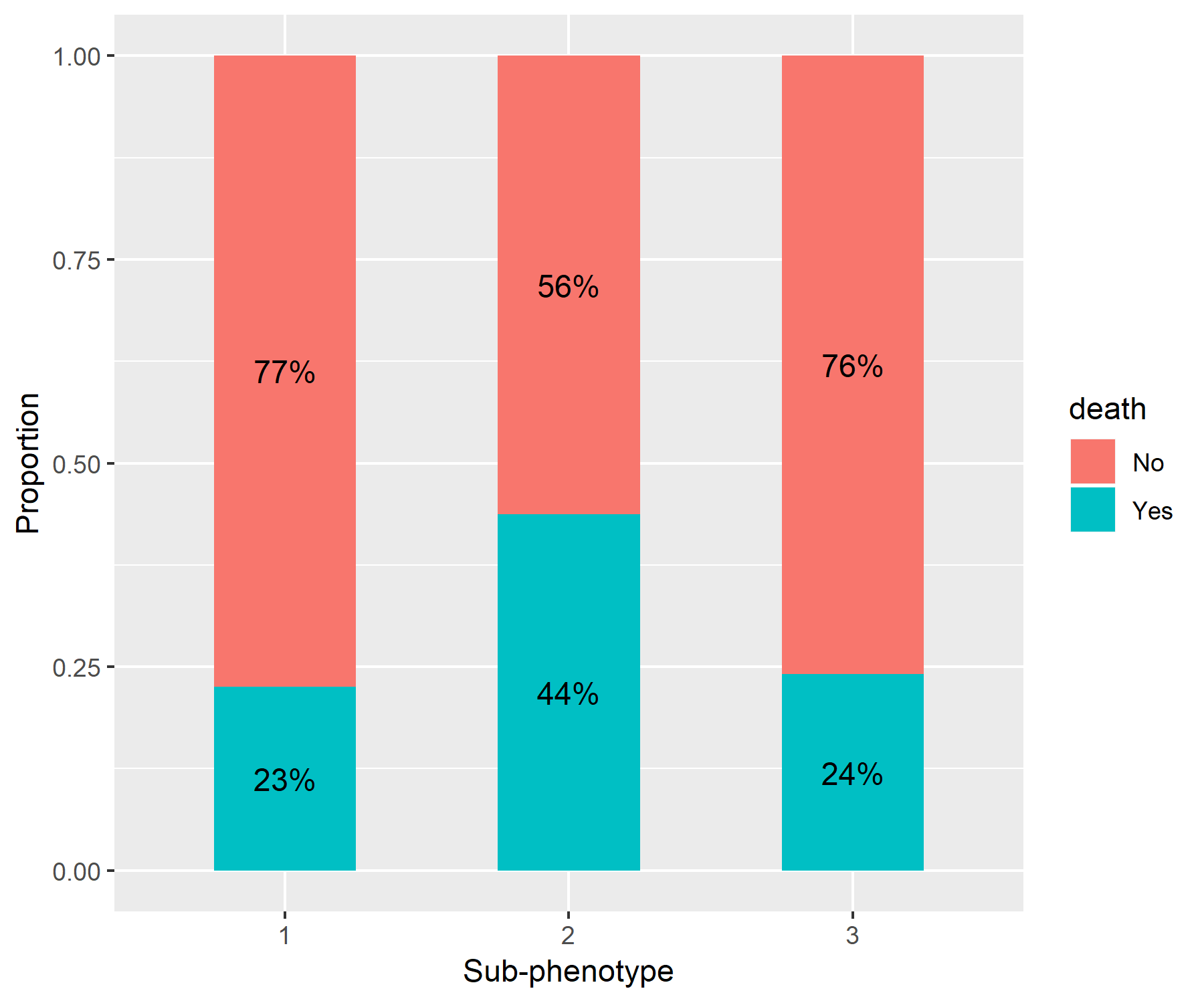}
\caption{Mortality by sub-phenotype.}
\label{fig-3}
\end{figure}

\subsection{Predictive Models}

Predictive models were built for all cases and each phenotype separately. Performances of predictive models evaluated using validation set are listed in Table \ref{table-2}.

Comparisons between models for phenotypes and for all cases indicate that significant improvements in AUC over existing predictive models are obtained for phenotype 1 and 2.  Phenotype 3 was found to be the most difficult to predict, with an AUC~90.9\% using GBM. However, even in this case, it is still a great improvement when compared to the LIPS score with a 81\% AUC \cite{gajic2011early}.

\begin{table}[t!]

\begin{center}
\begin{tabular}{l|l|l|l|l}
\hline
 \textbf{Phenotype} & \textbf{Metrics} & \textbf{GBM} & \textbf{Random Forest} & \textbf{P value} \\
\hline 

All & AUC & 0.904(0.897 - 0.911) & 0.904(0.897 - 0.911) & Reference \\
 &  Sensitivity & 0.905(0.837 - 0.931) & 0.9(0.88 - 0.916)  &  \\
  & Specificity & 0.765(0.745 - 0.825) & 0.782(0.768 - 0.799) & \\
  & Accuracy & 0.779(0.761 - 0.827) & 0.793(0.781 - 0.807)  &  \\
  & PPV & 0.289(0.275 - 0.338) & 0.303(0.291 - 0.317) &  \\
  & NPV & 0.987(0.98 - 0.99) & 0.987(0.984 - 0.989) &  \\
  \hline
1 &  AUC & 0.983(0.981 - 0.986) & 0.985(0.983 - 0.987) & $<$ 0.0001* \\
  & Sensitivity & 1(0.997 - 1) & 1(0.995 - 1)  &  \\
  & Specificity & 0.952(0.948 - 0.957) & 0.956(0.95 - 0.961)  &  \\
  & Accuracy & 0.953(0.949 - 0.957) & 0.956(0.951 - 0.961)  &  \\
  & PPV & 0.231(0.218 - 0.25) & 0.246(0.225 - 0.269)  &  \\
  & NPV & 1(1 - 1) & 1(1 - 1) \\
  \hline
2 &  AUC & 0.986(0.984 - 0.988) & 0.985(0.982 - 0.987) & $<$ 0.0001* \\
  & Sensitivity & 0.982(0.97 - 0.994) & 0.979(0.961 - 0.99) &   \\
  & Specificity & 0.957(0.953 - 0.961) & 0.955(0.938 - 0.96) &  \\
  & Accuracy & 0.957(0.953 - 0.961) & 0.956(0.939 - 0.961) &  \\
  & PPV & 0.391(0.368 - 0.414) & 0.382(0.31 - 0.411) &  \\
  & NPV & 0.999(0.999 - 1) & 0.999(0.999 - 1) &  \\
  \hline
3 &  AUC & 0.909(0.901 - 0.918) & 0.902(0.892 - 0.911) & 0.6538* \\
 &  Sensitivity & 0.902(0.829 - 0.93) & 0.879(0.826 - 0.919)  &  \\
  & Specificity & 0.769(0.749 - 0.843) & 0.773(0.733 - 0.82)  &  \\
  & Accuracy & 0.776(0.757 - 0.843) & 0.778(0.742 - 0.821)  &  \\
 &  PPV & 0.167(0.155 - 0.215) & 0.165(0.148 - 0.191)  &  \\
  & NPV & 0.994(0.99 - 0.995) & 0.992(0.989 - 0.994)  &  \\

\hline
\end{tabular}
\end{center}
\caption{\label{table-2} Performance of predictive models. *: AUC is compared with that obtained with all data (labeled as reference).}
\end{table}

\section{Discussion}

The study applied LCA to group ARDS patients into 3 sub-phenotypes and built separate predictive models for each phenotype. Using routinely available clinical variables, our LCA analysis identified three classes of ARDS that had different mortality rate, with sub phenotype 2 having significantly higher mortality (48\%) than the other 2 types.  Key characteristics of phenotype 2 include high bilirubin (liver failure), high creatinine/high pulse pressure (heart/renal failure), thrombocytopenia, lower minimum systolic blood pressure (hemodynamic instability), and lower PaO2/FiO2 indicating most severe respiratory failure. The identification of these subtypes may help triage ARDS patients that respond differently to treatment (e.g. fluids). At a AUC of .98 for this high mortality subgroup, our results indicate that significantly improved performance of prediction can be obtained for key ARDS sub-phenotypes.

It is known that heterogeneity in population poses a great challenge to predictive modeling. First, the training data may be comprised of instances from not just one distribution, but several distributions juxtaposed together. In the presence of multimodality within the classes, there may be imbalance among the distribution of different modes in the training set. Hence, some of the modes may be underrepresented during training, resulting in poor performance on those modes during the testing stage. Second, while some of the modes of a particular class may be easy to distinguish from modes of the other class, there may be modes that participate in class confusion, i.e., reside in regions of feature space that overlap with instances from other classes. The presence of such overlapping modes can degrade the learning of any classification model trained across all modes of every class. Third, even if we are able to learn a predictive model that shows reasonable performance on the training set, the test set may have a completely different distribution of data instances than the training set, as the populations of training and test sets can be different. Hence, the training performance can be quite misleading as it may not always be reflective of the performance on test instances. Due to these reasons, identifying homogeneous subgroups within heterogeneous population mitigate the impact of population’s heterogeneity in predictive model building.

A limitation of MIMIC III is that inflammatory or genetic biomarker data was not available. Studies indicates they may contribute in ARDS phenotyping\cite{calfee2014subphenotypes}. Combining clinical data and biological data in LCA-based phenotyping may improve homogeneities of ARDS sub-phenotypes, and further enhance predictive performance of machine learning within subgroups.

\section*{Acknowledgements}

Research reported in this publication was supported by a NIH SBIR award to CTA by NIH National Heart, Lung, and Blood Institute, of the National Institutes of Health under award number 1R43HL135909-01A1.

\makeatletter
\renewcommand{\@biblabel}[1]{\hfill #1.}
\makeatother

\bibliography{amia}

\appendix

\clearpage
\section{Appendix}

\begin{table}[h]
\tiny
\begin{center}
\begin{tabular}{l|l|l|l|l}
\hline
 \textbf{Identifier} & \textbf{Distribution} & \textbf{Volume} & \textbf{Shape} & \textbf{Orientation} \\
\hline 
EII & Spherical & equal & equal & NA  \\
VII & Spherical & variable & equal & NA  \\
EEI & Diagonal & equal & equal & coordinate axes  \\
VEI & Diagonal & variable & equal & coordinate axes  \\
EVI & Diagonal & equal & variable & coordinate axes  \\
VVI & Diagonal & variable & variable & coordinate axes  \\
EEE & Ellipsoidal & equal & equal & equal  \\
EEV & Ellipsoidal & equal & equal & variable  \\
VEV & Ellipsoidal & variable & equal & variable  \\
VVV & Ellipsoidal & variable & variable & variable  \\
 
\hline
\end{tabular}
\end{center}
\caption{\label{table-4} Models indexed in Figure \ref{fig-2}.}
\end{table}

\begin{table}[h]
\tiny
\begin{center}
\begin{tabular}{l|l|l|l|l}
\hline
 \textbf{Category} & \textbf{Name} & \textbf{Time Range} & \textbf{Features} & \textbf{Scale} \\
\hline 

Blood Chemistry & Anion gap & First 24 hours & Minimum, Mean, Maximum & Continuous \\
 & Albumin & First 24 hours & Minimum, Mean, Maximum & Continuous \\
 & Albumin & Across hospital stay & Minimum & Continuous \\
 & Bilirubin & First 24 hours & Minimum, Mean, Maximum & Continuous \\
 & Creatinine & First 24 hours & Minimum, Mean, Maximum & Continuous \\
 & Chloride & First 24 hours & Minimum, Mean, Maximum & Continuous \\
 & Glucose & First 24 hours & Minimum, Mean, Maximum & Continuous \\
 & Lactate & First 24 hours & Minimum, Mean, Maximum & Continuous \\
 & Potassium & First 24 hours & Minimum, Mean, Maximum & Continuous \\
 & Sodium & First 24 hours & Minimum, Mean, Maximum & Continuous \\
 & BUN & First 24 hours & Minimum, Mean, Maximum & Continuous \\
 & pH & Across hospital stay & Minimum & Continuous \\
 \hline
Blood Gas & SpO2 & First 24 hours & Minimum, Mean, Maximum & Continuous \\
 & PCO2 & First 24 hours & Minimum, Mean, Maximum & Continuous \\
 & PO2 & First 24 hours & Minimum, Mean, Maximum & Continuous \\
 & Bicarbonate & First 24 hours & Minimum, Mean, Maximum & Continuous \\
 & Tidal volume & First 24 hours & Minimum, Mean, Maximum & Continuous \\
 & Oxygen saturation & First 24 hours & Minimum, Mean, Maximum & Continuous \\
 & FIO2 & Across hospital stay & Maximum & Continuous \\
  \hline
Hematology & PTT & First 24 hours & Minimum, Mean, Maximum & Continuous \\
 & INR & First 24 hours & Minimum, Mean, Maximum & Continuous \\
 & PT & First 24 hours & Minimum, Mean, Maximum & Continuous \\
 & WBC & First 24 hours & Minimum, Mean, Maximum & Continuous \\
 & Hematocrit & First 24 hours & Minimum, Mean, Maximum & Continuous \\
 & Bands & First 24 hours & Minimum, Mean, Maximum & Continuous \\
 & Platelet & First 24 hours & Minimum, Mean, Maximum & Continuous \\
  \hline
Vital & Heart Rate & First 24 hours & Minimum, Mean, Maximum & Continuous \\
 & Systolic Blood Pressure & First 24 hours & Minimum, Mean, Maximum & Continuous \\
 & Diastolic Blood Pressure & First 24 hours & Minimum, Mean, Maximum & Continuous \\
 & Mean Airway Pressure & First 24 hours & Minimum, Mean, Maximum & Continuous \\
 & Respiratory Rate & First 24 hours & Minimum, Mean, Maximum & Continuous \\
 & Respiratory Rate & Across hospital stay & Maximum & Continuous \\
 & Temperature & First 24 hours & Minimum, Mean, Maximum & Continuous \\
  \hline
GCS & Total & First 24 hours & Minimum & Continuous \\
 & GCS motor & First 24 hours & Minimum & Continuous \\
 & GCS verbal & First 24 hours & Minimum & Continuous \\
 & GCS eye & First 24 hours & Minimum & Continuous \\
Output & Urine & Day 1, 2, 3 &  & Continuous \\
\hline
Flag & Alcohol abuse & &  & Binary(Yes/No) \\
 & Smoke &  &  & Binary(Yes/No) \\
 & Chemotherapy &  &  &  Binary(Yes/No) \\
 & Sepsis &  &  &  Binary(Yes/No) \\
 & Shock &  &  &  Binary(Yes/No) \\
 & Trauma &  &  &  Binary(Yes/No) \\
 & High risk surgery &  &  &  Binary(Yes/No) \\
 & Pneumonia &  &  &  Binary(Yes/No) \\
 & Smoke Inhalation &  &  &  Binary(Yes/No) \\
 & Near Drowning &  &  &  Binary(Yes/No) \\
 & Fracture &  &  &  Binary(Yes/No) \\
 & Diabetes &  &  &  Binary(Yes/No) \\
 & Admission: Urgent &  &  &  Binary(Yes/No) \\
 & Admission: Emergency &  &  &  Binary(Yes/No) \\
 & Admission: Selective &  &  &  Binary(Yes/No) \\
\hline
Demographic & Age &  &  &  Continuous \\
 & Gender &   &  & Binary (F/M) \\
 & BMI &   &  & Continuous \\

\hline
\end{tabular}
\end{center}
\caption{\label{table-3} Features used in predictive modeling.}
\end{table}

\end{document}